\documentclass{edm_template}
\usepackage{ifpdf}
\usepackage{cite}
\usepackage{wrapfig}
\usepackage{amsmath}
\usepackage{color}
\usepackage{algorithm}
\usepackage{algpseudocode}
\usepackage{setspace}
\let\Algorithm\algorithm
\renewcommand\algorithm[1][]{\Algorithm[#1]\setstretch{1.2}}
\usepackage{array}
\usepackage{mdwmath}
\usepackage{mdwtab}
\usepackage{eqparbox}
\usepackage{calc}
\usepackage{multirow}
\usepackage{graphicx}
\usepackage{fixltx2e}
\usepackage{bbm}
\usepackage[font={small}]{caption}
\usepackage{url}
\usepackage{subcaption}
\usepackage{tabularx}
\usepackage{fancyref}
\allowdisplaybreaks

\providecommand{\ie}{\emph{i.e.,} }

\providecommand{\parab}[1]{\noindent\textbf{#1}}

\begin{document}

\title{A Deep Learning Approach to \\ Behavior-Based Learner Modeling}
%
%
%
%
%

\numberofauthors{3} 
%
\author{
%
%
\alignauthor
Yuwei Tu\\
       \affaddr{Advanced Research}\\
       \affaddr{Zoomi Inc.}\\
       \email{yuwei.tu@zoomi.ai}
\alignauthor
Weiyu Chen\\
       \affaddr{Advanced Research}\\
       \affaddr{Zoomi Inc.}\\
       \email{weiyu.chen@zoomi.ai}
\alignauthor 
Christopher G. Brinton\\
       \affaddr{School of ECE}\\
       \affaddr{Purdue University}\\
       \email{cgb@purdue.edu}
}
\maketitle

\begin{abstract}
The increasing popularity of e-learning has created demand for improving online education through techniques such as predictive analytics and content recommendations. In this paper, we study learner outcome predictions, i.e., predictions of how they will perform at the end of a course. We propose a novel Two Branch Decision Network for performance prediction that incorporates two important factors: how learners progress through the course and how the content progresses through the course. We combine clickstream features which log every action the learner takes while learning, and textual features which are generated through pre-trained GloVe word embeddings. To assess the performance of our proposed network, we collect data from a short online course designed for corporate training and evaluate both neural network and non-neural network based algorithms on it. Our proposed algorithm achieves 95.7\% accuracy and 0.958 AUC score, which outperforms all other models. The results also indicate the combination of behavior features and text features are more predictive than behavior features only and neural network models are powerful in capturing the joint relationship between user behavior and course content.
\end{abstract}


\section{Introduction}
\label{sec:intro}

In the past decade, a variety of online learning platforms, like Massive Open Courses (MOOCs), have been drastically risen, offering educational services ranging from professional corporate training to higher education degrees. Predictive Learning Analytics (PLA) \cite{brinton2015mooc} is one of the most widely used techniques in the fields of learning analytics and educational data mining, giving instructors insights into how to predict educational student success and how to identify at-risk learners. The most two common sources of data for PLA are intermediate assessment results and users behavior data. For semester-long courses, frequent assessments are important to track student progress and the results of mid-term quizzes/assessments are predictive on final performance\cite{brinton2015mooc}; And for short-courses which do not have any intermediate assessment, users behavior data like their clickstreams on course content, are essential to final outcome prediction\cite{chen2017behavior}.

While PLA methods mostly focus on predicting students' knowledge stage, very few of them tried to involve course content into analysis. It has been demonstrated that course content data are also important in learning analytics, like identifying course prerequisite structure \cite{chenbehavioral}. Combining course content data and users' interaction on the corresponding course content can be the next step for providing more useful insights on course content selection and personalization. In this paper, we propose a new neural network based algorithm (defined and developed in Section \ref{sec:models}) which combines two aspects deemed important in determining course outcome: how learner behaves in the course and how the course content is constructed. 
Through evaluation (presented in Section \ref{sec:eval}) on the dataset that we collected from a short online course (explored in Section \ref{sec:data}), we find that our method successfully outperform other benchmarks, including both neural network based and non-neural network based algorithms. These results demonstrate that combining course content data and users' interaction on the corresponding course content can significantly improve the performance prediction accuracy.

\section{Related Work}
Approaches in PLA can be divided into three main categories: activity-based models, text-based models and network-based models.

\parab{Activity-based models.}
Activity-based models are the most common predictive models because activity data is the most abundant and granular data available from MOOC platforms. These methods use behavioral data, like clickstreams generated as the user interacts with content files, to predict both behaviour outcomes (i.e, dropout\cite{onah2014dropout}) and non-behaviour outcomes (i,e, quiz score\cite{brinton2015mooc, brinton2016mining}). The research of activity-based models starts with using general behavior frequencies. Overall, these studies found the frequency of user activities, is positively correlated to their final performance\cite{de2016role}. Additionally, some works attempt to capture the temporal nature of activity data by using more complex models to capture the user behavior patterns. These models, like Hidden Markov Model\cite{balakrishnan2013predicting}, high-order time series model\cite{brooks2015time} and Long Short-Term Memory neural network\cite{fei2015temporal}, are proved to be effective in predicting students' success, which also suggests that learners interests/intentions is time-varying and traceable.

\parab{Text-Based Models.}
Text-based models use natural language data generated by learners, like forum posts, notes, and comments on course content. Early researches on discussion forum data focused more on the social factors than the language itself, e.g., \cite{wen2014linguistic} which suggests that the number of posts or the average length of posts are more effective predictors of dropout for students than the text of posts itself. Moreover, there are several works which attempt to apply natural language processing to further understand the text features, but given the small sample of content within courses, most of them find that text-based features are less powerful than user-generated activity features\cite{robinson2016forecasting}.

\parab{Social Network-Based Models.}
Another family of prediction models are Social Learning Network (SLN)-based models, which infer the dynamics of learning behavior over a variety of graphs representing the relationships among the people and processes involved in learning\cite{brinton2014social}. Usually, threaded discussion forum is a prominent feature of every major MOOC platforms, so most researches use discussion fora to construct social networks where students are nodes and various reply relationships constitute edges.  

Overall, the above works have two main limitations: (a) only user activities are widely used as predictors, and (b) additional efforts are necessary to translate the results into actionable information. At present, most of the PLA methods are still focused on studying user behavioral patterns and user preference and seldom taking the course content into consideration. The method we develop in this paper, combining both textual features and user behavior features, outperforms other prediction algorithms with only user features. Additionally, our method shows potential of extracting the most significant course segments based on predictions and  translating prediction results into actionable course content selection.
\section{Features and Dataset}
\label{sec:data}
For model analytics and evaluation, we will use a dataset collected from an online course hosted by Zoomi Inc.\footnote{\url{https://zoomi.ai/}} The content of this course is a slideshow presentation consisting of 35 slides and 10 quiz questions; Each slide is considered to be a ``segment'' of content. Our dataset consists of the roughly 3,000 learners who enrolled in this course over a seven month period in 2017. 

We focus on two types of data captured by the platform: (i) clickstream data that logs each learner's navigation events with the slides, and (ii) course text data measured in words. Our methodology processes each of these data types into a set of features (explained in Sections 3.1\&2), and uses them in prediction models of learning outcomes, quantified as quiz performance (Section 3.3). In total, the dataset is comprised of roughly 900,000 clickstreams and 1,700 words across the video segments.

\subsection{User Behavioral Features}
Let $U$ denote the number of learners, indexed by $u \in \{1, 2, ..., U\}$, and let $S$ denote the number of segments, indexed by $s \in \{1, 2, \ldots, S\}$. The user behavioral features consists of several quantities for each segment $s$: time spent, view count, engagement, estimated time spent, and number of annotations:

\parab{Time Spent.} Time spent $t_{u,s}$ is the amount of (real) time learner $u$ spent on segment $s$. Formally, it is calculated as:
\begin{equation}
t_{u,s} = T_{u,s} - O_{u,s}
\label{eqn:time}
\end{equation}
where $T_{u,s}$ is the total recorded time learner $u$ spent on segment $s$, and $O_{u,s}$ is the subset of time for which the learner was off-task, e.g., with the application in the background \cite{chen2017behavior}.

\parab{View Count.} The view count $v_{u,s}$ is the number of times learner $u$ viewed segment $s$.

\parab{Expected Time Spent.} The expected time spent $\bar{t}_s$ is the time taken by users on segment $s$ on the average.

\parab{Notes/Highlights/Bookmarks.} $n_{u,s}$ is the number of annotations (notes, highlights, and/or bookmarks) that user $u$ makes on segment $s$.

\parab{Engagement.} Engagement $e_{u,s}$ quantifies the ``effort'' a learner has put into studying a segment. It is determined as:
\begin{equation}
e_{u,s}(t, b) = \min \left( \gamma \times \left ( \frac{1+ t / \bar{t}_s}{2} \right)^{\alpha_t} \left(\frac{1 + n / \bar{n}}{2} \right)^{\alpha_b} , 1 \right)
\label{eqn:eng}
\end{equation}
where $t$ and $n$ are the time spent and annotations made on the segment. Similar to $\bar{t}_s$, $\bar{b}$ represents the expected  number of annotations on a segment. $\alpha_a, \alpha_t \geq 0$ are parameters that allow diminishing marginal return on the corresponding variable. By doing so, a learner is rewarded (\ie higher engagement $e$) for distributing their activity more uniformly across segments \cite{chen2017behavior}. $\gamma \in (0,1]$ is an instructor-defined constant to control the engagement value.

With $S = 35$ different segments, this corresponds to $245$ features for each learner. Similar definitions of features have been seen to be predictive of user outcomes \cite{chen2017behavior} and engagement levels \cite{chenbehavioral} in previous research papers.

\begin{figure}[t]%
\centering
\includegraphics[width=0.52\textwidth]{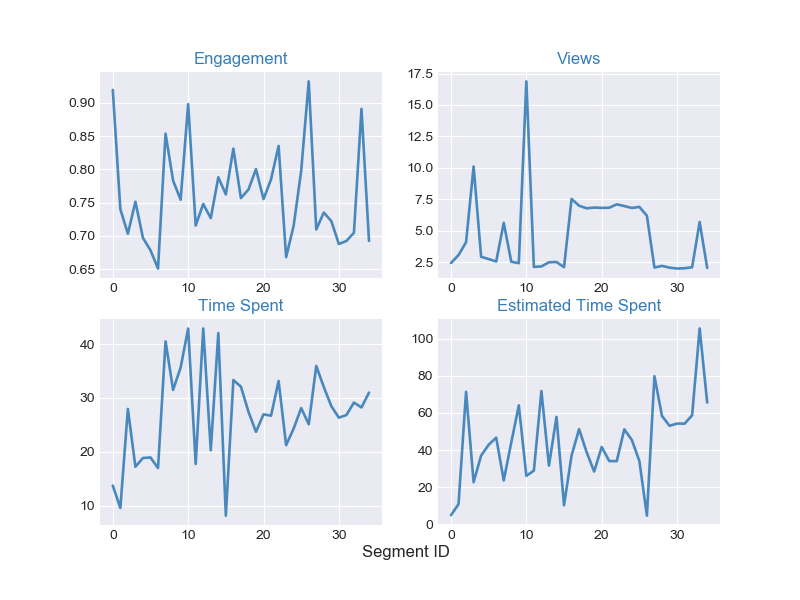}%
\caption{Average learner engagement, views, and time spent over segments, as well as estimated time spent in our dataset. There are clear cases of segments tending to receive significant focus as well as segments tending to be skipped over.}
\label{fig:behavior}%
\end{figure}

\parab{Descriptive Statistics.} Figure \ref{fig:behavior} shows the evolution of the average engagement, view count, time spent, and estimated time spent by segment; given there are more than 50\% missing values in the annotations, we do not include them here. Overall, we see that engagement and views have peaks and troughs throughout the course, showing that certain segments attract substantial focus while some also tend to be skipped over. This is different from e.g., Massive Open Online Courses (MOOCs) where participation tends to decline steadily over time, without any rapid change on particular pieces of content \cite{brinton2015mooc}. This indicates the potential of combining this information with content difficulty levels to identify segments that contribute most to final performance.

\subsection{Content Text Features}
We also leverage a set of text features that describe the topics of the course content. To obtain a textual representation of all segments, any multimedia content is first passed through open source speech-to-text conversion tools. Then, prior to feature processing, we perform the following steps for each segment's text:

\parab{Manual Correction and Noise Removal.} Translation errors may occur due to the nature of the speech-to-text conversion algorithm. Hence, we manually correct any translation errors and fill in any missing words. We also remove all punctuation and spaces, as they are not necessary in our line of text processing. Finally, we remove any text originating from headers and footers.

\parab{Tokenization, Lemmatization.} We perform tokenization to break down long strings of text to smaller units, i.e., tokens. For our task, we tokenize our text into a list of words.\footnote{ \url{https://nlp.stanford.edu/software/tokenizer.shtml}} We then normalize the tokens, which refers to transforming them to a consistent format, including converting all text to lower case letters and all numbers to their word equivalent. Finally, we perform lemmatization on every token to reduce the text variant forms to base form.\footnote{ \url{https://nlp.stanford.edu/IR-book/html/htmledition/tokenization-1.html}}

\parab{Stopword Removal.} After lemmatization, we remove stopwords according to a standard, aggressive list.\footnote{ \url{http://www.lextek.com/manuals/onix/stopwords1.html}}

\parab{Topic Modeling.} Because of the limited amount of textual data (less than 2,000 words), applying standard natural language processing (NLP) techniques such as Latent Dirichlet Allocation (LDA) and TF-IDF may not be optimal. Instead, we resort to GloVe embeddings, which are pre-trained word embeddings based on web-scale data\footnote{\url{https://nlp.stanford.edu/projects/glove/}}. GloVe has demonstrated high performances in several NLP tasks, such as part-of-speech tagging, text segmentation \cite{tu2018educational}, machine translation. GloVe focuses on the co-occurrence probabilities $P(i|j)$ between two words, i.e., how often word $i$ appears given word $j$ in a context \cite{pennington2014glove}. We use embedding vectors of dimension 100 trained on the Wikipedia 2014 and Gigaword datasets. Higher dimensions were seen to increase the potential of overfitting in our proposed neural network model of Section 4.

\subsection{Outcome Variable}
\begin{figure}[t]%
\centering
\includegraphics[width=0.5\textwidth]{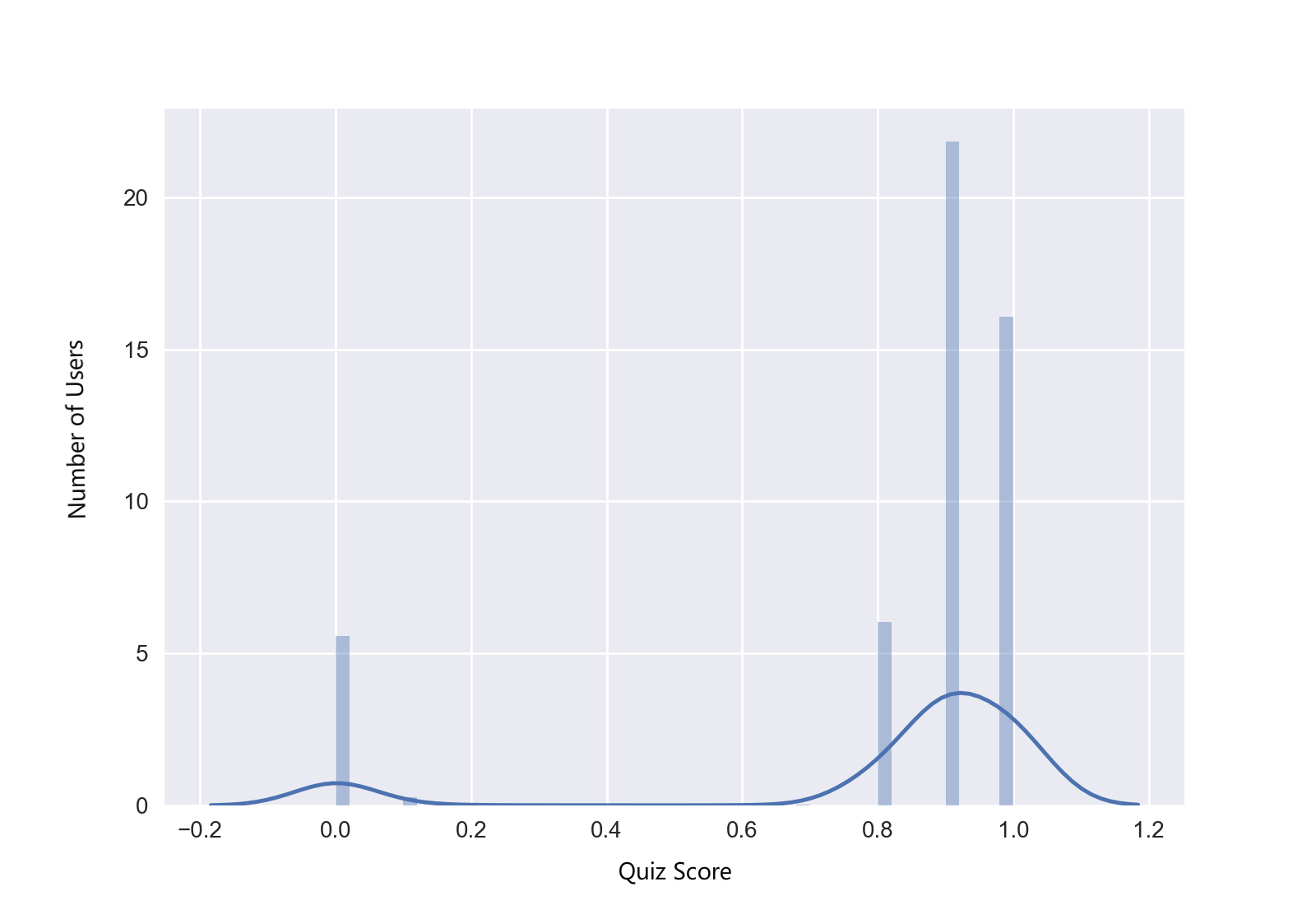}%
\caption{User performance data: majority of users were able to achieve preferred outcome in the course without reattempting the quiz. }
\label{fig:performance}%
\end{figure}

We treat user performance on quiz questions as the outcome variable. Letting $P_u$ be learner $u$'s final score in the course, with each of the 10 questions counting for $0.1$ points, $P_{u} \in \{0, 0.1, 0.2, ..., 1\}$. In Figure \ref{fig:performance}, we plot a histogram of the performance distribution. Users tend to spread on both ends of the distribution, either with a high passing score above 0.8, or with a relatively low score. Hence, we further partition users into two groups: pass or fail, with 0.8 being the cutoff. Note that since a heavy majority of users receive passing outcomes, the dataset is imbalanced and a classifier may obtain greater than 90\% accuracy by simply predicting all pass. In Section 5, we will discuss how we cope with class imbalance via data re-sampling, stratified cross validation, choice of metric, and model penalty parameters.  

\section{Models}
\label{sec:models}
In this section, we propose three deep learning models that leverage the features from Section 4: the Baseline, which is a typical one-layer fully connected neural network model; the Embedding Similarity Network, which weights the measurements of each segment with the text similarity between content segments and quiz questions; and the Two Branch Decision Network, where the cosine similarity is replaced with another fully connected layer.

\begin{figure}[hbt!]
	\centering
	\begin{subfigure}{.4\linewidth}
	\centering
	\includegraphics[width=1.0\linewidth]{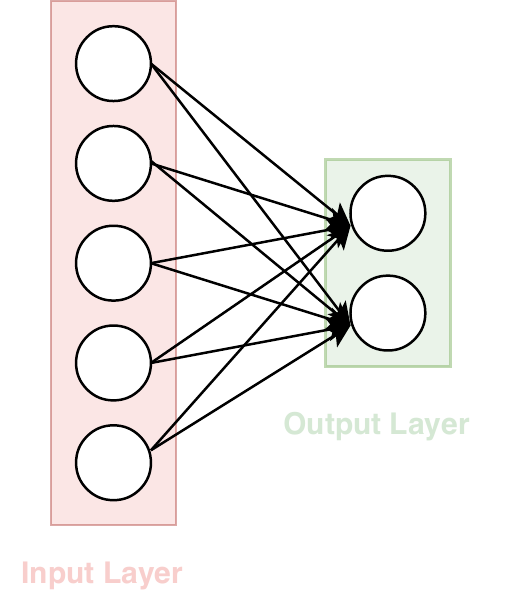}
	\subcaption{Baseline Network}
	\label{fig:first}
	\end{subfigure}
	
    \qquad
	\begin{subfigure}{.95\linewidth}
	\centering
	\includegraphics[width=1.0\linewidth]{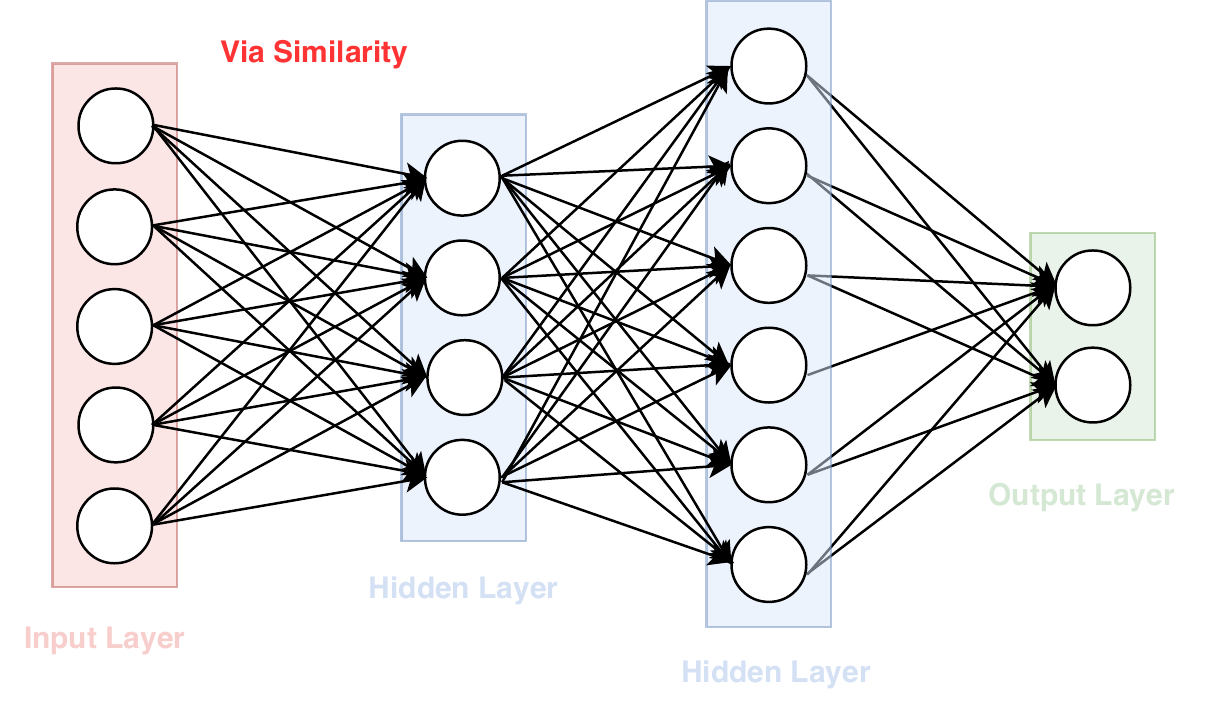}
	\subcaption{Embedding Similarity Neural Network}
	\label{fig:second}
	\end{subfigure}
	
	\begin{subfigure}{.95\linewidth}
	\centering
	\includegraphics[width=1.0\linewidth]{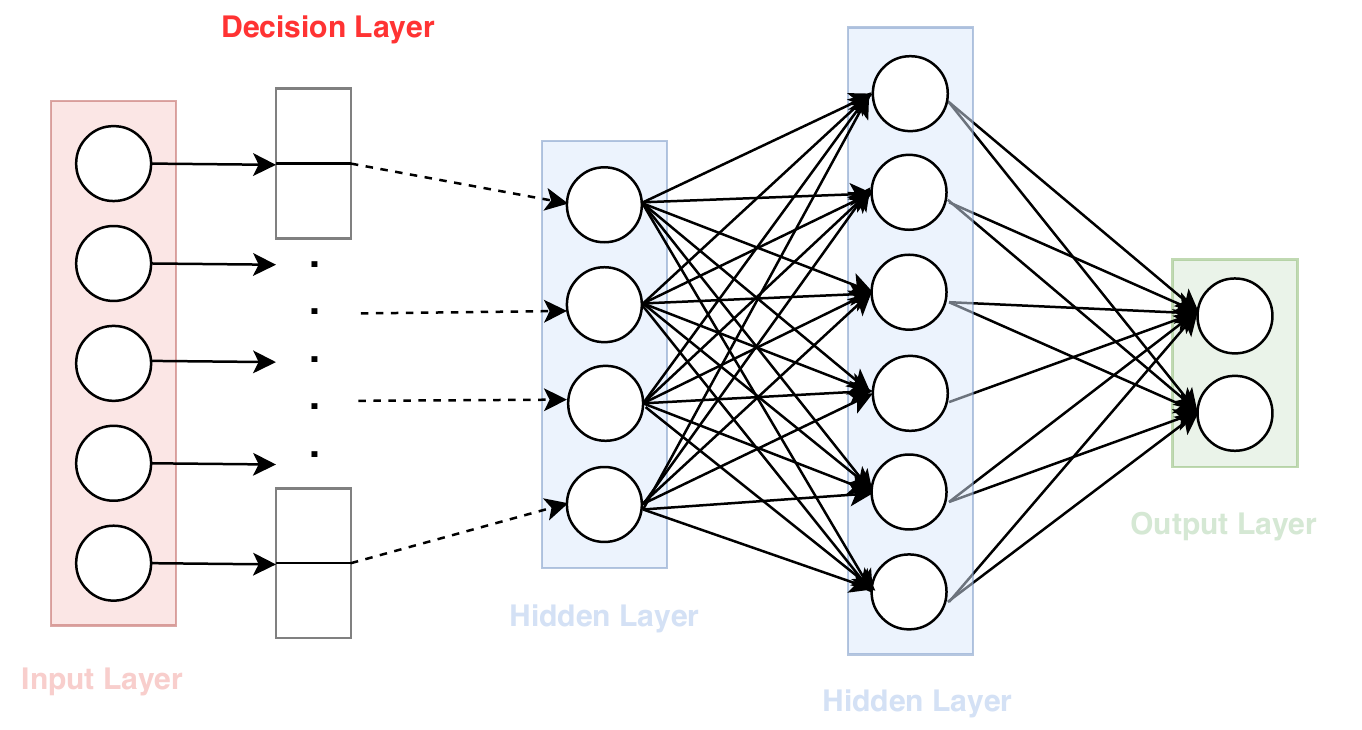}
	\centering
	\subcaption{Two Branch Decision Network}
	\label{fig:third}
	\end{subfigure}
	
	\caption{(a) demonstrates the one layer fully-connected network. (b) demonstrates a one layer fully-connected network built based on pre-selected features based on text similarities. (c) demonstrates a two branch decision neural network, whereas the two branch textual input decides the input of the behavioral features.} 
\label{fig:network}
\end{figure}

\subsection{Baseline: One-layer Fully Connected Network}
We first derive a one-layer fully connected neural network for the purpose of baseline comparison. The reason we chose a neural network instead of other classifiers, such as Random Forest, is to provide a fair comparison of our proposed model in terms of structure and model performance.  An one-layer fully connected network is sufficient in our case as the dataset has 2,914 learners with 245 features to consider. We want to increase the ratio of number of data points to number of features to provide enough training information for network tuning.

Denoting the feature vector for learner $u$ as $x_u$, the model is as follows:
\begin{equation}
\label{model:baseline_network}
\centering
P_{u} = \sigma (\mathbf{W} x_{u} + b)
\end{equation}
Here, the matrix $\mathbf{W}$ contains the weights for matrix multiplication, and $b$ is a bias vector. The choice of activation function $\sigma$ is softmax. This model is visualized in Figure 3(a).

\subsection{Embedding Similarity Network}
Now we proceed to our first proposed Embedding Similarity Network. The intuition behind the Embedding Similarity Network is to choose the input $x_{u,d} \in x_u$ based on how informative the input feature variable $x_{u,d}$ might be. We first assign text similarity value based on the following the cosine similarity the corresponding $T_c$ and the $T_q$. The more relevant one specific segment is to the quiz, the more likely the $x_{u,d}$ that takes place in this segment is indicative of the user outcome $P_u$.

Figure \ref{fig:network}(b) visualizes the network structure, and Algorithm 1 formalizes the algorithm. Depending on the similarity value, a subset of the input layer $\mathcal{X}$ is passed to the first hidden layer. Finally, a fully connected layer is used calculate the final class logits, that eventually determine the prediction output. 

\begin{algorithm}[t]
\begin{algorithmic}[1]
\State $\textbf{Initialize}$ network weights and biases with truncated normal distribution 
\For{Each $x_{u,d}$ in input network layer $\mathcal{X}$,}
	\State Find corresponding content segment $T_c$
	\State Calculate similarity $\frac{T_c \dot \sum_{q} T_q}{|T_c| \dot |\sum_{q} T_q|}$, for $q \in [1, 2, ..., 10]$
	\State Pass $x_{u,d}$ to the hidden layer $h$ if similarity $>$ threshold
\EndFor
\State logits = softmax($h \times weights + bias$)
\State $\textbf{return}$ logits 
\end{algorithmic}
\caption{Embedding Similarity Network}
\label{pseudo:embedding}
\end{algorithm}

\subsection{Two Branch Decision Network}
Thirdly, we propose a two branch decision network, visualized in Figure \ref{fig:network}(c) and formalized in Algorithm 2. This network exploits textual relationship between the content and quiz text. Instead of feeding all user behavioral data directly, our proposed network uses a two branch decision structure before feeding in all user behavioral data. In particular, for every segment, the corresponding segment content is compared against all the quiz content via a fully connected layer.

\begin{algorithm}[t]
\begin{algorithmic}[1]
\State $\textbf{Initialize}$ network weights and biases with truncated normal distribution 
\For{Each $x_{u,d}$ in input network layer $\mathcal{X}$,}
	\State Extract content segment $T_c$
	\State Extract summed $\sum_{q} T_q$, for $q \in [1, 2, ..., 10]$
	\State  $x_{u,d}$ * binary($(T_c * \sum_{q} T_q) \times weights + bias $)
	\State Pass to the hidden layer $h$ 
\EndFor
\State logits = softmax($h \times weights + bias$)
\State $\textbf{return}$ logits 
\end{algorithmic}
\caption{Two Branch Decision Network}
\label{pseudo:two_branch}
\end{algorithm}

\section{Experiments}
\label{sec:eval}

\subsection{Classifiers and Procedure}
We now consider several choices of classifiers in our dataset: the baseline neural network (BNN), the embedding similarity network (ESN), and finally the two branch decision network (TBN). We also investigated other approaches, such as recurrent neural networks, but found sub-optimal performances on our dataset. We additionally considered a baseline using a non-neural network based Gradient Boosting Classifier (GBC) for comparison. The gradient boosting classifiers have demonstrated high performance in predicting student outcomes in other works, e.g., \cite{superby2006determination,chen2017behavior}. 

TBN has 10x more parameters when compared to BNN, due to the number of layers and the number of neurons in each layer. Given a limited number of training samples, a large number of parameters in training require methods such as dropout rate \cite{srivastava2014dropout}or residual network \cite{szegedy2017inception} to reduce the likelihood of over-fitting. 

\parab{Initialization.} For each network-based method, all parameters are initialized according to the following rules. All weights are initialized following a truncated normal distribution with a zero mean and a zero bias. As our data is normalized, the expectation of network parameters should remain zero or close to zero. Nonetheless, zero initialization of weights under optimization perform very poorly. On the other hand, initialization with very small numbers also reduce efficiency in optimization, as smaller numbers take a longer time to converge. Hence, we consider a truncated normal distribution for weight initialization to be appropriate in our scenario.

\parab{Metric.} We primarily consider AUC (\ie the area under the ROC curve) \cite{huang2005using} and accuracy in our evaluation metric. In practice, we are dealing with an imbalanced dataset that has only a few number of users who failed the course. Similarly to email spam prediction, using accuracy as evaluation metric would not suffice, as predicting all pass would produce a `high' accuracy. Consequently, we seek a classifier to obtain a high accuracy with a high AUC score at the same time. Moreover, for the network-based methods (TBN, ESN, and BNN), we also inspect the cross-entropy. In ideal scenarios, our classifiers should produce labels with high probability, \ie high confidence level.

\parab{Parameter Settings and Estimation.} The network's fully connected layer is set at a dimension of $8$. For training and evaluation, we divide the dataset into $K$ stratified folds ($K = 5$) such that each fold has the same proportion of each class, and during training, the neural network is trained by minimizing cross entropy via the Adam Optimizer with a learning rate of $0.005$. For each training epoch, we feed a mini-batch of 50 data points to the network for reducing internal covariate shift in training data \cite{ioffe2015batch}. All network training ends after 2000 epochs. Afterwards, we may vary these parameters individually to further optimize one single network.

\begin{figure*}[t]
\centering
\begin{subfigure}{.32\linewidth}
\label{fig:first}
\centering
\includegraphics[width=\textwidth,]{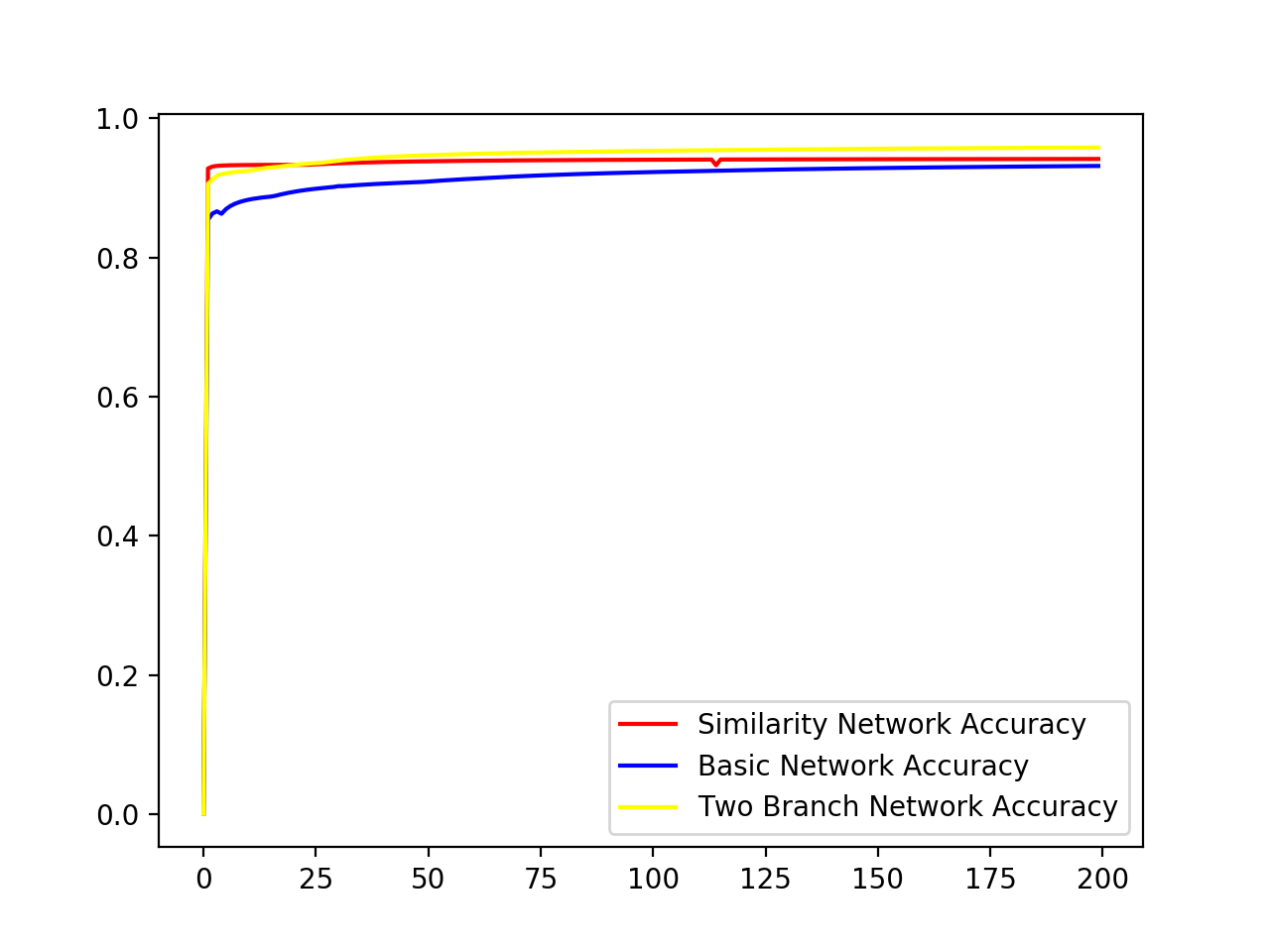}
\subcaption{Accuracy} 
\end{subfigure}
\begin{subfigure}{.32\linewidth}
\label{fig:second}
\centering
\includegraphics[width=\textwidth]{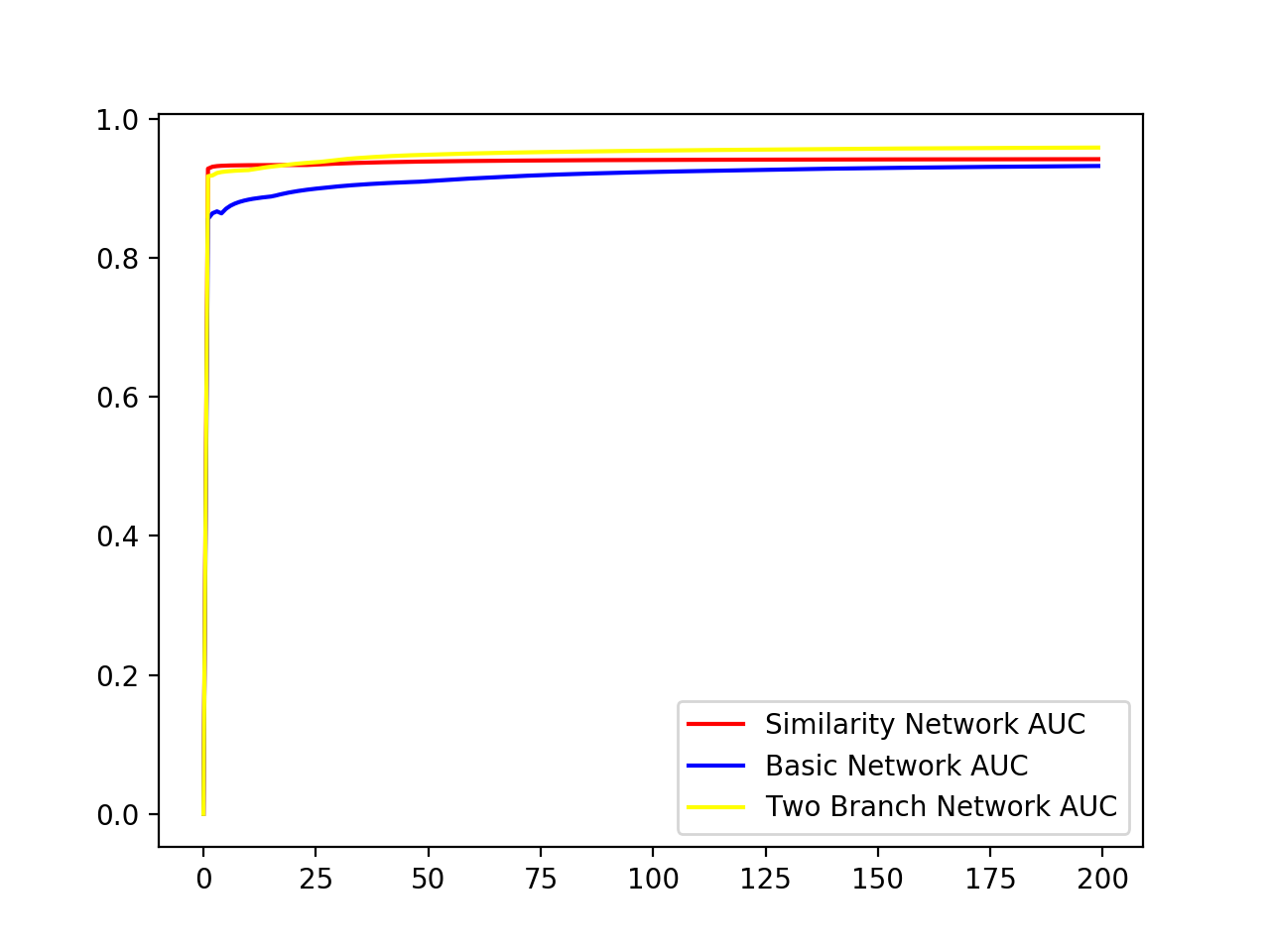}
\subcaption{AUC} 
\end{subfigure}
\begin{subfigure}{.32\linewidth}
\label{fig:third}
\centering
\includegraphics[width=\textwidth]{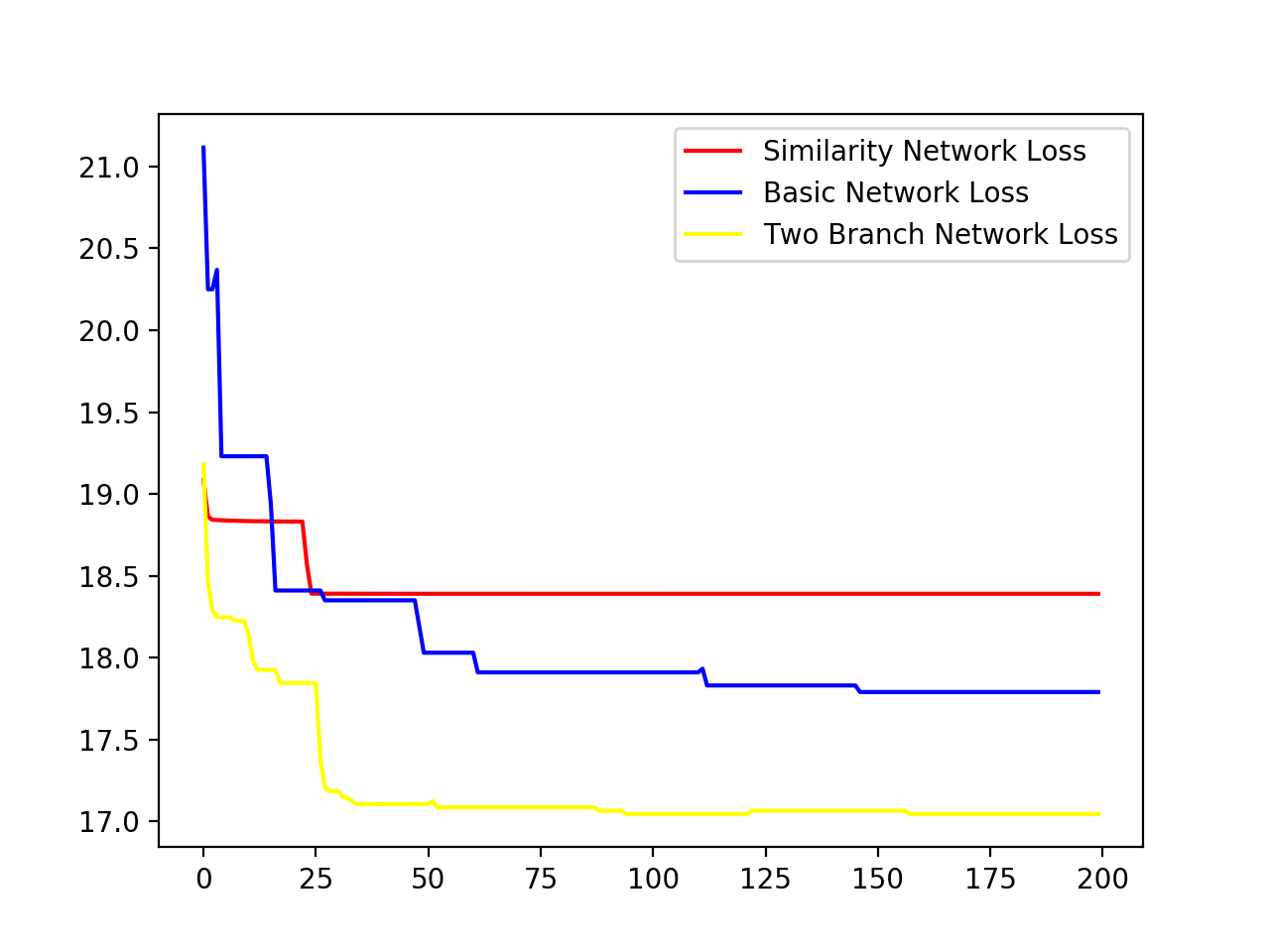}
\subcaption{Loss} 
\end{subfigure}
\caption{(a) and (b) show increasing accuracy and AUC scores for all three neural network models over training iterations. (c) demonstrates the reduction on the cross entropy loss against number of training epochs.}
\label{fig:performances}
\end{figure*}

\subsection{Model Evaluation}
The results are tabulated in Table \ref{tab: performance}, and plotted over iterations for each neural network algorithm in Figure 4.

Overall, from Table 1, we see that all four models achieve prediction accuracy greater than 90\% and AUC greater than 0.9, which demonstrates that the user behavioral features are highly informative in the prediction of a learner's final performance. Compared to the performance of our proposed neural-network models and the gradient boosting model, even the gradient boosting model achieves the highest accuracy, but considering AUC is fairly often preferred over accuracy for binary classification (particularly when there is class imbalance), our TBN model presents the highest overall prediction quality, which demonstrates neural network models are more powerful and stable in the quiz prediction tasks.

Among the three proposed neural network models (BNN, ESN and TBN), both ESN and TBN achieve higher accuracy and AUC than BNN, which proves our previous assumption that the combination of text-based features and user behavioral features is more powerful than behavioral features alone. The TBN performs even better than ESN, which proves that it is advisable to use a fully connected layer rather than simply calculating cosine similarity to model the joint relationship between course content and quiz questions. 

Overall, our Two Branch Network achieves the highest AUC among all other models including the gradient boosting classifier. This demonstrates the importance of incorporating course content into performance prediction and inspires further research on selecting text features based on performance prediction results. As indicated in Figure \ref{fig:performances}, the AUC/accuracy curves reach their peaks and then smoothen after 200 epochs which indicates a further improvement of early stopping conditions.

\begin{table}[t]
\centering
\begin{tabular}{llll}
\hline
Algorithm          & Accuracy & AUC   & Cross Entropy \\ \hline
BNN   & 0.932    & 0.932 & 17.79         \\
ESN & 0.942    & 0.942 & 18.39         \\
TBN & 0.957    & 0.958 & 17.04         \\
GBC     & 0.976    & 0.911 & --           \\ \hline
\end{tabular}
\caption{Prediction performance of the neural-network and non-network-based models, for each of the three metrics. All cases obtain AUCs of greater than $0.9$.}
\label{tab: performance}
\end{table}

\section{Conclusion}
\label{sec:conclusion}
 In our paper, we propose a novel Two Branch Decision Network for performance prediction that incorporates two types of educational data: how learners progress through the course, and how the content progresses through the course. We combine clickstream features that log every action the learner takes while learning and textual features generated through pre-trained GloVe word embeddings. To evaluate the performance of our proposed network, we collected the data from a short online course which contains 35 slides and 10 quiz questions, and tested both neural network and non-neural network based algorithms. The results demonstrate that our neural-network model achieves relatively better performance than non-network models, and that incorporating text features into the prediction model can significantly improve the prediction performance over training on clickstream data alone. 
 
 In the future, our continued research will focus on expanding the current feature space, like incorporating users interactive records with finer granularity, or incorporating text features of course content with a predefined knowledge graph; and improving the current network connections into more advanced structures, such as a Recurrent Neural Network (RNN), to capture the underlying sequence of course content. Moreover, we will expand our experiment targets from corporate training courses to MOOCs, where the course structure is more complicated and the user behavior pattern is more flexible\cite{brinton2015mooc}.

\bibliographystyle{abbrv}
\bibliography{main}
\end{document}